\documentclass[journal]{IEEEtran}

\usepackage{graphicx}
\usepackage{amsmath}
\usepackage{times}  
\usepackage{helvet}
\usepackage{courier}
\usepackage[hyphens]{url} 
\usepackage{graphicx} 
\usepackage{subfigure}
\usepackage{amsfonts}
\usepackage{color}
\usepackage[]{booktabs}
\usepackage[switch]{lineno}
\usepackage[pagebackref=false,breaklinks=true,letterpaper=true,colorlinks,bookmarks=false]{hyperref}
\usepackage{cite}
\usepackage{float}

\ifCLASSINFOpdf
\else
\fi

\begin{document}
\title{Weakly-supervised 3D Human Pose Estimation with Cross-view U-shaped Graph Convolutional Network}

\author{Guoliang Hua$^{*}$,
        Hong Liu$^{\dagger}$,
        Wenhao Li$^{*}$,
        Qian Zhang,
        Runwei Ding, and
        Xin Xu
        
\thanks{$^{*}$ Equal contribution.}
\thanks{$^{\dagger}$ Corresponding author.}
\thanks{G. Hua, H. Liu, W. Li, Q. Zhang, and R. Ding are with the Key Laboratory of Machine Perception,
Peking University Shenzhen Graduate School, Shenzhen 518055, China. 
E-mail: \{glhua, hongliu, wenhaoli, qian.zhang, dingrunwei\}@pku.edu.cn. 

Xin Xu is with the College of Intelligence Science and Technology, National University of Defense Technology, Changsha, 410073, China. E-mail: xinxu@nudt.edu.cn.

This work is supported by National Natural Science Foundation of China (No.62073004, No.61825305),
Shenzhen Fundamental Research Program (No. GXWD20201231165807007-20200807164903001, JCYJ20190808182209321, JCYJ20200109140410340). 
}
}

\markboth{}
{Hua \MakeLowercase{\textit{et al.}}: Weakly-supervised 3D Human Pose Estimation with Cross-view U-shaped Graph Convolutional Network}

\maketitle

\begin{abstract}
Although monocular 3D human pose estimation methods have made significant progress, it is far from being solved due to the inherent depth ambiguity. Instead, exploiting multi-view information is a practical way to achieve absolute 3D human pose estimation. In this paper, we propose a simple yet effective pipeline for weakly-supervised cross-view 3D human pose estimation. By only using two camera views, our method can achieve state-of-the-art performance in a weakly-supervised manner, requiring no 3D ground truth but only 2D annotations. Specifically, our method contains two steps: triangulation and refinement. First, given the 2D keypoints that can be obtained through any classic 2D detection methods, triangulation is performed across two views to lift the 2D keypoints into coarse 3D poses. Then, a novel cross-view U-shaped graph convolutional network (CV-UGCN), which can explore the spatial configurations and cross-view correlations, is designed to refine the coarse 3D poses. In particular, the refinement progress is achieved through weakly-supervised learning, in which geometric and structure-aware consistency checks are performed. We evaluate our method on the standard benchmark dataset, Human3.6M. The Mean Per Joint Position Error on the benchmark dataset is 27.4 mm, which outperforms existing state-of-the-art methods remarkably (27.4 mm vs 30.2 mm). 
\end{abstract}


\IEEEpeerreviewmaketitle

\section{Introduction}
\IEEEPARstart{3D} HUMAN pose estimation aims to produce a 3-dimensional figure that describes the spatial position of the depicted person. This task has drawn tremendous attention in the past decades~\cite{li20143d,chen20173d,zhou2017towards}, playing a significant role in many applications such as action recognition, virtual and augmented reality, human-robot interaction, etc. Many recent works~\cite{martinez2017simple,fabbri2020compressed,strided,mhformer} focus on estimating 3D human poses from monocular inputs, either images or 2D keypoints. However, it is ill-posed due to the inherent depth ambiguity since multiple 3D poses can map to the same 2D keypoints. As a result, most monocular methods only estimate the relative positions to the root joint and fail to estimate the absolute 3D poses, which greatly limits practical applications. Instead, exploiting multi-view information is arguably the best way to achieve absolute 3D pose estimation~\cite{harvesting2017}. 

\begin{figure}[tb]
  \centering
  \subfigure[Volumetric approach]{\includegraphics[width=0.71\linewidth]{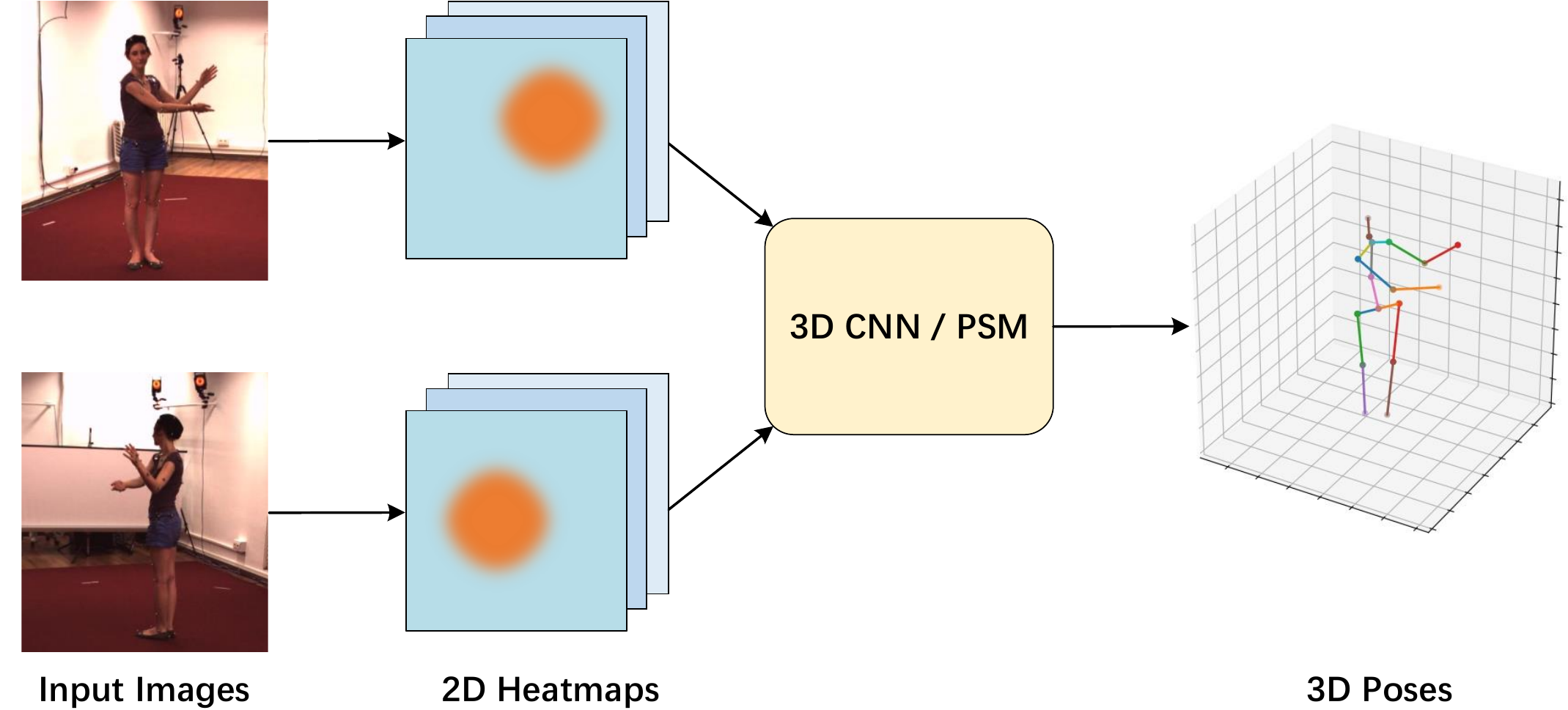}}
  \label{fig:pipline1}
  \subfigure[Cross-view refinement (ours)]{\includegraphics[width=1.0\linewidth]{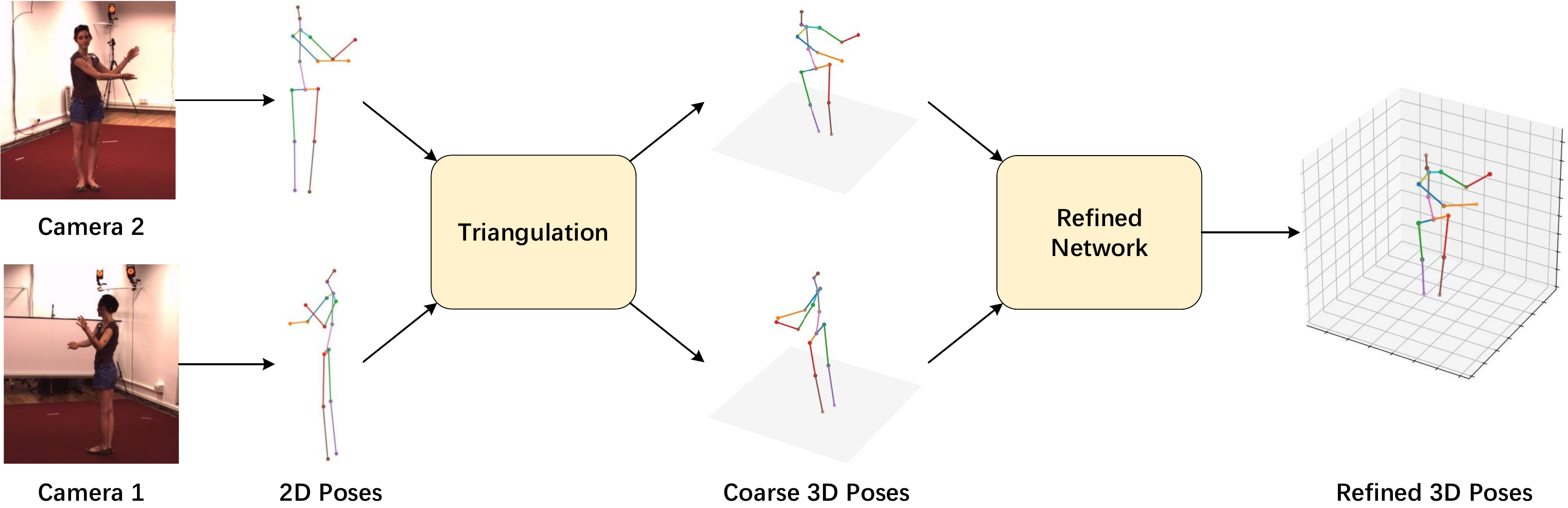}}
  \label{fig:pipline2}
  \caption
  {
    (a) Most state-of-the-arts use multi-view images as input and follow a pipeline that estimates 2D heatmaps and then directly recovers 3D poses through volumetric convolutional neural networks or Pictorial Structure Model (PSM). 
    (b) Instead, we consider a 2D-3D lifting pipeline in a coarse-to-fine manner, which first obtains coarse 3D poses through triangulation from cross-view 2D joint detections and then refines the pose with a refinement model.
  }
  \label{fig:pipeline}
\end{figure}

\begin{figure*}[htb]
  \centering
  \includegraphics[width=1 \linewidth]{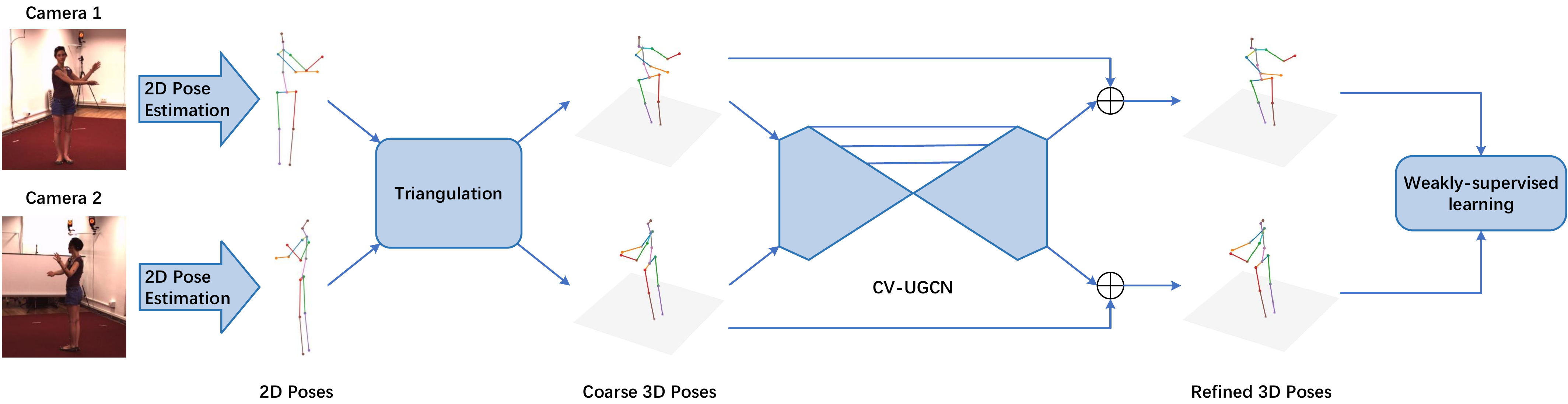}
  \caption
  {
      The proposed pipeline for cross-view weakly-supervised 3D human pose estimation. 
      Given the 2D poses estimated from RGB images of two different camera views, the triangulation is first performed to obtain coarse 3D poses. 
      Then, a cross-view U-shaped graph network (CV-UGCN) trained in a weakly-supervised manner is used to produce realistic and structurally plausible 3D poses. 
  }
  \label{fig:overview}
\end{figure*}
	
Multi-view human pose estimation methods benefit from the complementary information from different camera views, e.g. multi-view geometric constraints to resolve the depth ambiguity and different views of the depicted person to deal with the occlusion problem. Many existing multi-view based methods~\cite{iskakov2019learnable,qiu2019cross,kocabas2019self} follow a pipeline that first takes multi-view images as input to predict 2D detection heatmaps and then projects them to 3D poses through volumetric convolutional networks or Pictorial Structure Model (PSM)~\cite{harvesting2017,chen2014articulated}, as shown in Figure~\ref{fig:pipeline} (a). However, using the convolutional neural network to perform 2D-3D lifting requires quantities of labelled 3D data as supervision, which is difficult and costly to collect. PSM discretizes the space around the root joint by an $N \times N \times N$ grid and assigns each joint to one of the $N^3$ bins (hypotheses), therefore requiring no 3D ground truth. However, the 2D-3D lifting accuracy of PSM based method is subject to the number of grids, and the computation complexity is of the order of $O(N^6)$ which is computationally expensive.

To release the requirement of large quantities of 3D ground truth and take the computation complexity into consideration, a simple yet effective pipeline (see Figure~\ref{fig:pipeline} (b)) is proposed in this paper for cross-view 3D human pose estimation. Different from existing methods, our method estimates 3D human poses from coarse to fine and contains two steps: triangulation and refinement.  Considering the increasing number of camera views will bring more computation and reduce the flexibility of application in the wild, we only use two camera views for training and inference. In the first step, we perform the triangulation between two camera views to lift 2D poses, which can be obtained through any classic 2D keypoint detection methods, to the 3D space. 
However, the triangulated 3D poses are noisy and unreliable due to the errors of 2D keypoint detection and camera parameters calibration, thus requiring further refinement.

In the refinement progress, a lightweight cross-view U-shaped graph convolutional network (CV-UGCN) is designed to refine the coarse 3D poses. 
By taking the cross-view coarse 3D poses as input, CV-UGCN is able to exploit spatial configurations and cross-view correlations to refine the poses to be more reasonable. 
Meanwhile, CV-UGCN is trained in a weakly-supervised manner, requiring no 3D ground truth but only 2D annotations. Specifically, by making full use of the cross-view geometric constraints, geometric and structure-aware consistency checks are introduced as the learning objective to train the network end-to-end. 

We summarize our contributions as follows:
\begin{itemize}
    \item 
    A simple yet effective pipeline is proposed for cross-view 3D human pose estimation, which estimates the 3D human poses from coarse to fine by using the triangulation and the refinement model. 
    \item 
    A cross-view U-shaped graph convolutional network (CV-UGCN), which can take advantage of spatial configurations and cross-view correlations, is proposed as the refinement model.
    
    \item 
    A weakly-supervised learning objective containing geometric and structure-aware consistency checks is introduced, therefore releasing from the requirement of large quantities of 3D ground truth for training. 
\end{itemize}

Extensive experiments have been conducted on the benchmark dataset, Human3.6M, to verify the effectiveness of our method. The Mean Per Joint Position Error (MPJPE) on the benchmark dataset is 27.4 mm, which outperforms existing state-of-the-art methods remarkably (27.4 mm vs 30.2 mm). 

\section{Related Work}
\noindent \textbf{Single-view 3D pose estimation.} 
Current promising solutions for monocular 3D pose estimation can be divided into two categories. 
Methods of the first category directly regress the 3D poses from monocular images. 
Pavlakos \emph{et al.}~\cite{pavlakos2017coarse} introduced a volumetric representation for 3D human poses, while requiring a sophisticating deep network architecture that is impractical in application. 
In the second category, these works first estimate 2D keypoints and then lift 2D poses to the 3D space (2D-3D lifting).  
Martinez \emph{et al.}~\cite{martinez2017simple} predicted 3D poses via a fully-connected residual network and showed low error rates when using 2D ground truth as input. 
Cai \emph{et al.}~\cite{cai2019exploiting} presented a local-to-global GCN to exploit spatial-temporal relationships to estimate 3D poses from a sequence of skeletons. 
Meanwhile, they introduced a pose refinement step to further improve the estimation accuracy. 
However, they only utilized the 2D detections to constrain the depth-normalized poses, while ignoring the refinement for depth values. Different from~\cite{cai2019exploiting}, we perform both 3D transformation and 2D reprojection consistency checks in our refinement model, so that the refinement is more sufficient.

\begin{figure*}[t]
	\centering
	\includegraphics[width=1.0 \linewidth]{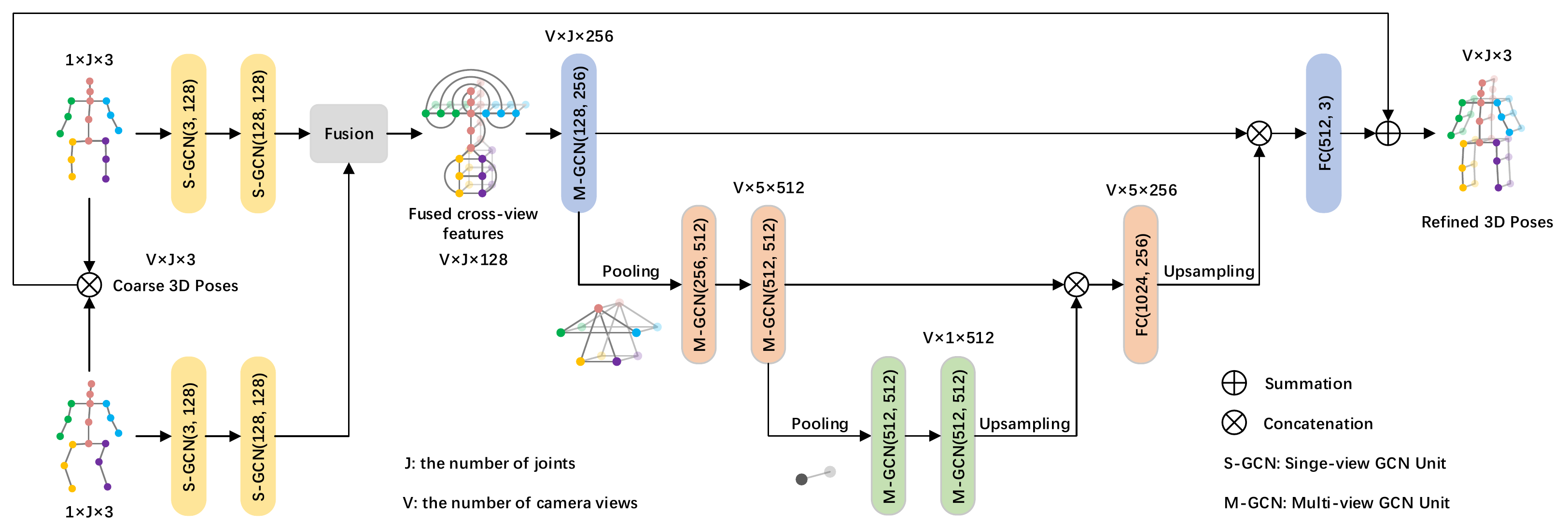}
	\caption
	{
		A schematic of CV-UGCN. S-GCN units are first utilized to preprocess the coarse 3D poses of each view to capture spatial configurations independently. 
        Then, the fused cross-view features are fed into the U-shaped architecture, where M-GCN units are utilized to explore additional cross-view correlations. 
  	}
	\label{fig:network}
\end{figure*}

\noindent \textbf{Multi-view 3D pose estimation.}
In order to estimate the absolute 3D poses, recent works seek to utilize information from multiple synchronized cameras to solve the problem of depth ambiguity. 
Most multi-view based approaches use 3D volumes to aggregate 2D heatmap predictions.
Qiu \emph{et al.}~\cite{qiu2019cross} presented a cross-view fusion scheme to estimate 2D heatmaps of multiple views and then used a recursive Pictorial Structure Model to estimate the absolute 3D poses.
Iskakov \emph{et al.}~\cite{iskakov2019learnable} proposed a learnable triangulation method to regress 3D poses from multiple views. However, volumetric approaches are computationally demanding. 
To recover 3D poses from multi-view images without using compute-intensive volumetric grids, Remelli \emph{et al.}~\cite{remelli2020lightweight} exploited 3D geometry to fuse input images into a unified latent representation of poses. 
Different from these methods that embedded the improved 2D detector into their model to obtain more accurate 2D poses to further improve the 3D pose estimation, our method focuses on the task of 2D-3D lifting and can be easily integrated with any 2D detectors to achieve 3D pose estimation with a lightweight refinement model.  

\noindent \textbf{Weakly/self-supervised methods.}
Because 3D human pose datasets are limited and collecting 3D human pose annotations is costly, researchers have resorted to weakly or self-supervised approaches. 
Zhou \emph{et al.}~\cite{zhou2017towards} proposed a weakly-supervised transfer learning method for in-the-wild images.
RepNet~\cite{wandt2019repnet} proposed a weakly-supervised reprojection network by using an adversarial training approach. 
Moreover, in~\cite{kundu2020self}, a self-supervised learning method was proposed to estimate 3D poses from unlabeled video frames via part guided human image synthesis.
Compared with previous methods, our method has the advantage of decomposing the challenging 3D human pose estimation task into two steps and making full use of geometric and structure-aware consistency checks for weakly-supervised learning.

\section{Cross-view 3D Human Pose Estimation}
Figure~\ref{fig:overview} depicts our pipeline for weakly-supervised cross-view 3D human pose estimation. Given the estimated 2D poses ${x}_{i} \in \mathbb{R}^{J \times 2}$ from two different views, we aim at recovering their absolute 3D poses ${X}_{i} \in \mathbb{R}^{J \times 3}$, where $i$ is the index of the camera views, and $J$ is the number of joints. 
In particular, we first reconstruct coarse 3D poses through triangulation. Then, a cross-view U-shaped graph convolutional network (CV-UGCN) is proposed to refine the coarse triangulated 3D poses to obtain more precise estimations.

\subsection{Triangulation}
Assuming two cameras are synchronized and calibrated, triangulation can be performed between two camera views to lift 2D poses into the 3D space. Given the 2D joint locations $x_1$, $x_2$ of two camera views, which can be obtained through classic 2D keypoint detection methods, the triangulation is solved through:
\begin{equation}
\left[\begin{array}{l}
	x_{1}^{j} \times \left(T_{c_1, w} \cdot \tilde{X}_{w}^{j}\right) \\
	x_{2}^{j} \times \left(T_{c_2, w} \cdot \tilde{X}_{w}^{j}\right)
\end{array}\right]  = 0,
\end{equation}
where $x_1^j$, $x_2^j$ are the 2D coordinates of $j$-th joint, $T_{c_1, w}$, $T_{c_2, w}$ are the transformation matrixes between the camera $c_i$ and world coordinate system. Since the origin of the world coordinate system can be set in any positions, we select one of the cameras as the origin to simplify the computation. Then, the triangulation is defined as: 
\begin{equation}
\label{eq:trig1}
\left[\begin{array}{l}
x_{1}^{j} \times \left(I \cdot \tilde{X}_{1}^{j} \right)\\
x_{2}^{j} \times \left(T_{c_2, c_1} \cdot \tilde{X}_{1}^{j}\right)
\end{array}\right]  = 0,
\end{equation}
where $I$ is the identity matrix, and camera $c_1$ is set as the origin of the world coordinate system. By solving Eq.~(\ref{eq:trig1}) through Singular Value Decomposition (SVD), we can obtain the 3D pose $\tilde{X}_1$ of the camera $c_1$. Similarly, we can set $c_2$ as the origin to obtain the 3D pose $\tilde{X}_2$ of the camera $c_2$.

Although triangulation is a straightforward way to achieve 2D-3D lifting, it is subject to the accuracy of the 2D joint detections and the precision of calibrated camera parameters. To solve this problem, Qiu \emph{et al.} proposed a more robust Recursive Pictorial Structure Model (RPSM) to replace the triangulation~\cite{qiu2019cross}. Different from them, we propose to optimize the initial triangulated 3D poses through a weakly-supervised learning refinement model, which is lightweight and requires no 3D annotations to train. 

\begin{figure*}[t]
	\centering
	\includegraphics[width=0.8 \linewidth]{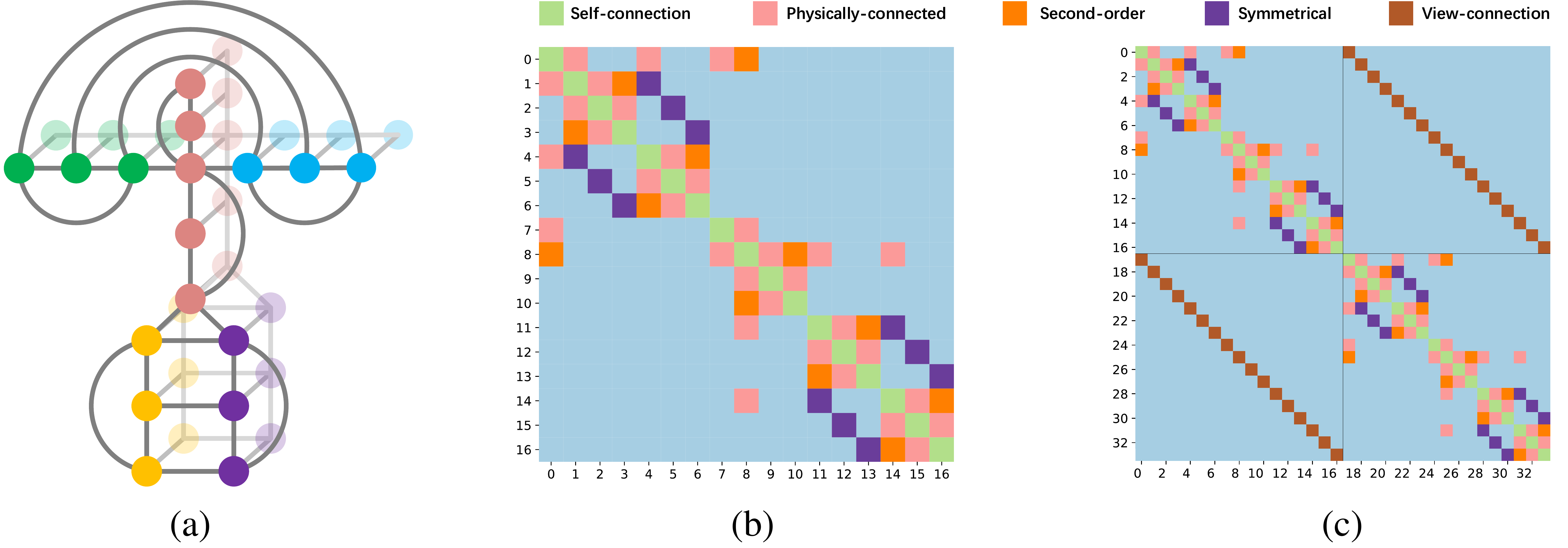}
	\caption
	{
		(a) The human skeleton graph in kinematic connections and cross-view connections. 
		(b) The adjacency matrix of single-view GCN.
		(c) The adjacency matrix of multi-view GCN (2 views).
	}
	\label{fig:graph}
\end{figure*}

\subsection{Cross-view Refinement Model}
In order to refine the coarse triangulated 3D poses, a cross-view refinement model is proposed. Specifically, we design a cross-view U-shaped graph convolutional network, named CV-UGCN, which can make full use of the spatial configurations and cross-view correlations for refinement. As specified in Figure~\ref{fig:network}, the CV-UGCN takes the cross-view coarse 3D poses as input and outputs the residual shift added to the coarse 3D poses to obtain the refined 3D poses. 

\subsubsection{Graph modeling}
Here, we first give the definition of the graph model of CV-UGCN. 
The cross-view skeletons are organized as an undirected graph in the spatial and camera-view domains. 
The undirected graph $\mathbb{G}$ contains a set of vertices $\mathcal{V}$ and edges $\mathcal{E}$, 
where $\mathcal{V}=\left\{v_{i j} \mid i=1, \ldots, V ; j=1, \ldots, J\right\}$ corresponds to $J$ joints of the human body in $V$ camera views. 
For single-view GCN (S-GCN) that processes data from a single view, the edge $\mathcal{E}$ only consists of kinematic connections of spatial configurations. 
For multi-view GCN (M-GCN) that embeds additional cross-view correlations into the graph model, the edge $\mathcal{E}$ consists of two parts: kinematic connections and cross-view connections.

\subsubsection{Graph convolution}
As presented in Figure~\ref{fig:graph}, the defined undirected graph is represented in an adjacency matrix $A \in \mathbb{R}^{N \times N}$, where $N = VJ $ ($N = J$ for S-GCN). 
In detail, the graph nodes are classified as the neighboring nodes according to their semantic meanings in the human body structure, and five kernels are used for different neighboring nodes: 
(i) self-connection nodes;
(ii) physical-connection nodes; 
(iii) second-order connection nodes; 
(iv) symmetrical nodes; 
(v) view-connection nodes. 
Note that S-GCN only processes single-view data, thus having no view-connection nodes. 

Given the input signal $H \in \mathbb{R}^{N \times C}$ with $C$ channels, following~\cite{cai2019exploiting}, we update the graph convolution operation in~\cite{kipf2016semi} by dismantling adjacent matrix into $k$ sub-matrices to:
\begin{equation}
  \label{eq:G-CONV 2}
  \begin{array}{ll}
      Z& =  \mathcal{C}\left( H, A; W\right)\\
       & = \sum_{k} \tilde{D_{k}}^{-\frac{1}{2}} \tilde{A_{k}} \tilde{D_{k}}^{-\frac{1}{2}} H W_{k},
  \end{array}
\end{equation}
where $Z \in \mathbb{R}^{N \times F}$ is the convolved signal matrix, $\mathcal{C}$ is the function of the graph convolution. $W_{k} \in \mathbb{R}^{C \times F}$ and $\tilde{A_{k}} \in \mathbb{R}^{N \times N}$ are the learnable filter matrix and the normalized adjacency matrix for the $k$-th type of neighboring nodes respectively, and $\tilde{D_{k}}^{i i}=\sum_{j} \tilde{A_{k}}^{i j}$.

\subsubsection{Network structure} 
As specified in Figure~\ref{fig:network}, S-GCN and M-GCN units are the basic blocks to build our CV-UGCN. 

\noindent \textbf{Single-view GCN module. }
The triangulated results exist a lot of noises that may disturb the accurate cross-view information interaction.
Therefore, the cross-view coarse 3D poses are first fed into S-GCN units independently to capture the spatial configurations. Exploiting the adjacent matrix $A^s$ of single-view GCN shown in Figure~\ref{fig:graph}, the operation of S-GCN units is defined as:
\begin{equation}
  \label{eq:SGCN}
      F^{s}_{i} = SGCN\left( \tilde{X}_{i}\right) = \mathcal{C}\left( \tilde{X}_{i}, A^s; W^s\right),
\end{equation}

In this way, the coarse poses of each view are mapped into a high-dimension feature space, and the structure-aware information can be fully explored in each view.

\noindent \textbf{Multi-view U-GCN module. }
After the spatial configurations are fully explored in each view independently, we concatenate them into a shared latent space via the fusion layer to fuse the features of two views for cross-view information interaction, 
\begin{equation}
  \label{eq:fusion}
      F^{f} =  F^s_1 \otimes F^s_2,
\end{equation}
where $\otimes$ represents the concatenation along view dimension. 
The fused features are then input into M-GCN units so that the cross views can complement each other. Besides the spatial connections, view connections are considered in the adjacent matrix $A^m$ of M-GCN. The operation of M-GCN units is defined as:
\begin{equation}
  \label{eq:MGCN}
      F^{m} =  MGCN\left( F^f\right) = \mathcal{C}\left( F^f, A^m; W^m\right).
\end{equation}
In M-GCN, both the spatial configurations and cross-view correlations are extracted. 

Furthermore, inspired by the success of U-shaped architecture~\cite{ronneberger2015u,cai2019exploiting,wang2020motion}, we build a UGCN architecture by exploiting the M-GCN units, graph pooling, and graph upsampling operations, as shown in Figure~\ref{fig:network}. As a result, the features in the cross-view spatial graph model can be integrated in a multi-scale manner, assisting in obtaining more accurate pose refinement.

\noindent \textbf{Refinement. }
Thanks to the spatial configurations and cross-view correlations that are fully explored in our CV-UGCN, the precise residual shift is predicted for 3D pose refinement. By adding the residual shift to the triangulated 3D coarse poses $\tilde{X}$, we can obtain the accurate refined poses:
\begin{equation}
  X = \mathcal{G}(\tilde{X})+\tilde{X},
\end{equation}
where $\mathcal{G}$ denotes the refinement function of CV-UGCN. 

\section{Weakly-supervised Learning Approach}
To relieve the requirement of 3D ground truth for supervised learning, a weakly-supervised learning objective is designed and only needs 2D annotations that are much easier to gather. Geometric and structure-aware consistency checks are performed in both single-view and cross-view manners. 

\subsection{Single-view Optimization}
By considering the 2D reprojection consistency and left-right symmetry priors, we first impose the geometric and structure-aware consistency checks for each view to perform single-view optimization.

\subsubsection{2D reprojection loss} 
After the coarse 3D pose is refined by CV-UGCN, we reproject the refined 3D pose into the 2D joint location through:
\begin{equation}
    x = \frac{1}{Z} \cdot K \cdot X,
\end{equation}
where $X$ is the refined 3D pose, $x$ is the reprojected 2D pose, $K$ is the camera intrinsic parameter, and $Z$ is the depth value. Then, the 2D reprojection loss is introduced to penalize the error between $x$ and 2D ground truth $y$:
\begin{equation}
  L_{r}= \sum_{i=1}^{V} \sum_{j=1}^{J}\left\|{y}_{i}^{j}-{{x}}_{i}^{j}\right\|_{2},
\end{equation}
where $i$ and $j$ index the camera views and joints, respectively.

\subsubsection{Symmetry loss} Noticing that the structure of the human body is left-right symmetrical, a symmetry loss is introduced to constrain the bone length of the left part and the right part to be the same. It is effective to alleviate the depth ambiguity problem,  especially when the occlusion happens~\cite{dabral2018learning}. The symmetry loss is defined as: 
\begin{equation}
  L_{s}=\sum_{i=1}^{V} \sum_{k}\left\|{B}_{i}^{k}-{{B}}_{i}^{C(k)}\right\|_{2},
\end{equation}
where ${B}_{i}^{k} = \left\|{X}_{i}^{k_{1}}-{X}_{i}^{k_2}\right\|$ is the estimated bone length for a left-side bone $k$ of $i$-th camera view, and $C(k)$ is the corresponding right-side bone, $k_{1}$ and $k_{2}$ are the joints of the $k$-th bone.

\begin{figure}[t]
  \centering
  \includegraphics[width=1 \linewidth]{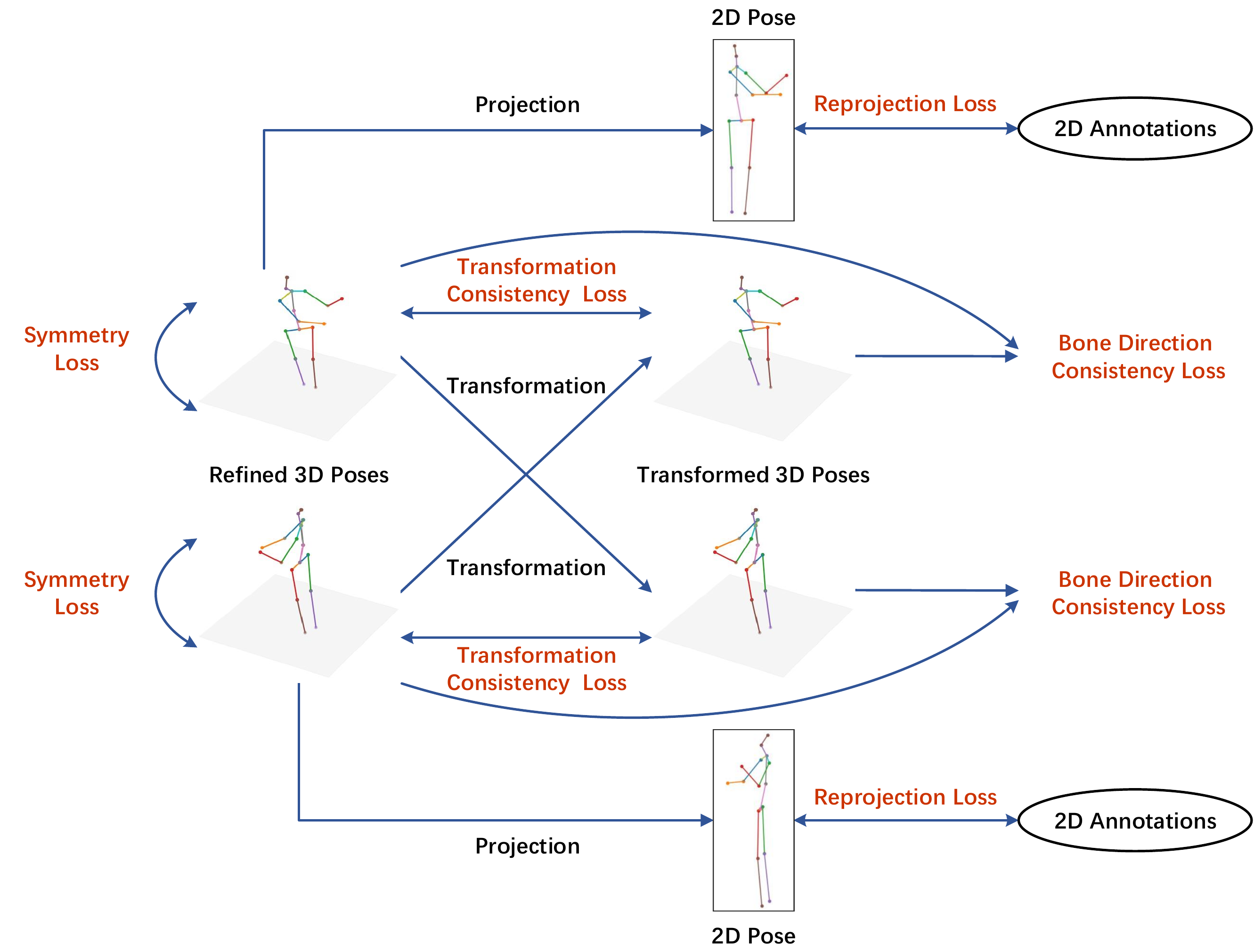}
  \caption
  {
    Weakly-supervised learning objective. 
    Geometric and structure-aware consistency checks are performed in a single view and cross views to train the refinement model in a weakly-supervised manner. 
  }
  \label{fig:loss}
\end{figure}  

\subsection{Cross-view Optimization}
It is quite often that some joints are unseen or close to each other in one view, which brings significant challenges for single-view optimization. Therefore, we perform cross-view optimization to take advantage of cross-view consistency priors. 

Given the refined 3D poses of two views, ${X}_{1}, {X}_{2}$, cross-view 3D transformation is performed to generate the poses of camera view $c_1$ from $c_2$ by:
\begin{equation}
    {X}_{1\leftarrow 2} = T_{c_1,c_2} \cdot {X}_{2}.
\end{equation}
Similarly, we can obtain the ${X}_{2\leftarrow 1}$ transformed from ${X}_{1}$.

\subsubsection{3D transformation consistency loss}
The transformed poses should keep consistency with the original 3D poses, thus we calculate the 3D transformation consistency loss as follows:
\begin{equation}
  \small
  L_{t}= 
  \sum_{i=1}^{V} \sum_{j=1}^{J} (\left\|{X}_{1}^{j}-{X}_{1\leftarrow 2}^{j}\right\|_{2} + \left\|{X}_{2}^{j}-{X}_{2\leftarrow 1}^{j}\right\|_{2}).
\end{equation}

\begin{table*}[t]
  \centering
    \caption
    {
    Quantitative comparisons on Human3.6M dataset under protocol \#1.
    3D means that the method uses 3D annotations to train the model. 
    $V$ is the number of camera views for training and inference. 
    The best score is marked in bold, second best is underlined.
    }
  \resizebox{\textwidth}{!}{
  \begin{tabular}{@{}l|cc|ccccccccccccccc|c@{}}
  \toprule
  \textbf{Protocol \#1} &$V$ &3D & Dir. & Disc & Eat & Greet & Phone & Photo & Pose & Purch. & Sit & SitD. & Smoke & Wait & WalkD. & Walk & WalkT. & Avg.\\
  \midrule
  
  Wandt \emph{et al.} (CVPR'19)~\cite{wandt2019repnet}& 1& & 77.5& 85.2& 82.7& 93.8& 93.9& 101.0& 82.9& 102.6& 100.5& 125.8& 88.0& 84.8& 72.6& 78.8& 79.0& 89.9 \\
  
  Tome \emph{et al.} (CVPR'17)~\cite{tome2017lifting}& 1& & 65.0& 73.5& 76.8& 86.4& 86.3& 110.7& 68.9& 74.5& 110.2& 173.9& 85.0& 85.6& 86.2& 71.4& 73.1& 88.4 \\
  
  Wang \emph{et al.} (ICCV'19)~\cite{wang2019distill}&1 & &- &- &- &- &- &- &- &- &- &- &- &- &- &- &- &86.4\\
  
  
Xu \emph{et al.} (CVPR'21)~\cite{xu2021graph} &1 &\checkmark &45.2 &49.9 &47.5 &50.9 &54.9 &66.1 &48.5 &46.3 &59.7 &71.5 &51.4 &48.6 &53.9 &39.9 &44.1 &51.9 \\

  
  
  
  Pavllo \emph{et al.} (CVPR'19)~\cite{pavllo20193d}&1 &\checkmark & 45.2 & 46.7 & 43.3 & 45.6 & 48.1 & 55.1 & 44.6 & 44.3 & 57.3 & 65.8 & 47.1 & 44.0 & 49.0 & 32.8 & 33.9 & 46.8 \\
  

  Liu \emph{et al.} (CVPR'20)~\cite{liu2020attention}&1 &\checkmark &41.8 &44.8 &41.1 &44.9 &47.4 &54.1 &43.4 &42.2 &56.2 &63.6 &45.3 &43.5 &45.3 &31.3 &32.2 &45.1 \\

  Zeng \emph{et al.} (ECCV'20)~\cite{zeng2020srnet}&1 &\checkmark& 46.6& 47.1& 43.9& 41.6& 45.8& 49.6& 46.5& 40.0& 53.4& 61.1& 46.1& 42.6& 43.1& 31.5& 32.6& 44.8 \\


  Chen \emph{et al.} (TCSVT'21)~\cite{chen2021anatomy}&1 &\checkmark &41.4 &43.5 &40.1 &42.9 &46.6 &51.9 &41.7 &42.3 &53.9 &60.2 &45.4 &41.7 &46.0 &31.5 &32.7 &44.1 \\
  
  Li \emph{et al.} (TMM'22)~\cite{strided} &1 &\checkmark &{40.3} &{43.3} &{40.2} &{42.3} &{45.6} &{52.3} &{41.8} &{40.5} &{55.9} &{60.6} &{44.2} &{43.0} &{44.2} &{30.0} &{30.2} &{43.7} \\
  
  \midrule
  
  Trumble \emph{et al.} (BMVC'17)~\cite{trumble2017total}&4 &\checkmark & 92.7& 85.9& 72.3& 93.2& 86.2& 101.2& 75.1& 78.0& 83.5& 94.8& 85.8& 82.0& 114.6& 94.9& 79.7& 87.3\\
  
  Kundu \emph{et al.} (CVPR'20)~\cite{kundu2020self}&4 & &- &- &- &- &- &- &- &- &- &- &- &- &- &- &- &85.8\\
  
  Iqbal \emph{et al.} (CVPR'20) ~\cite{iqbal2020weakly}&4 & &- &- &- &- &- &- &- &- &- &- &- &- &- &- &- &67.4\\
  
  
  Pavlako \emph{et al.} (CVPR'17) ~\cite{harvesting2017}&4 &\checkmark &41.2 &49.2 &42.8 &43.4 &55.6 &46.9 &40.3 &63.7 &97.6 &119.0 &52.1 &42.7 &51.9 &41.8 &39.4 & 56.9 \\
  
  Tome \emph{et al.} (3DV'18) ~\cite{tome2018rethinking}&4 & &43.3& 49.6& 42.0 &48.8 &51.1 &64.3 &40.3 &43.3 &66.0 &95.2 &50.2 &52.2 &51.1 &43.9 &45.3 &  52.8 \\
  
  Chen \emph{et al.} (CVPR'19)~\cite{chen2019weakly}&2 & & 41.1& 44.2& 44.9& 45.9& 46.5& 39.3& 41.6& 54.8& 73.2& 46.2& 48.7& 42.1& 35.8& 46.6& 38.5& 46.3 \\

  Qiu \emph{et al.} (ICCV'19)~\cite{qiu2019cross}&4 & &28.9 &32.5 &26.6 &28.1 &\textbf{28.3} &\textbf{29.3} &28.0 &36.8 &41.0 &\textbf{30.5} &35.6 &30.0 &28.3 &30.0 &30.5 &31.2 \\

  He \emph{et al.} (CVPR'20)~\cite{he2020epipolar}&4 & &29.0 &\underline{30.6} &27.4 &\underline{26.4} &31.0 &31.8 &\underline{26.4} &\underline{28.7} &\underline{34.2} &42.6 &32.4 &\underline{29.3} &\underline{27.0} &\underline{29.3} &\underline{25.9} &30.4 \\
  
  Remelli \emph{et al.} (CVPR'20)~\cite{remelli2020lightweight}&4 &\checkmark & \underline{27.3} &32.1 & \textbf{25.0} &26.5 &29.3 &35.4 & 28.8 &31.6 &36.4 &\underline{31.7} &\underline{31.2} &29.9 &\textbf{26.9} &33.7 &30.4 &\underline{30.2} \\
  
  \midrule
  Ours&2 & &\textbf{25.6}&\textbf{27.7} &\underline{25.3} &\textbf{24.5} &\underline{29.1} &\underline{29.5} &\textbf{23.5} &\textbf{25.7} &\textbf{31.2} &37.3 &\textbf{28.9} &\textbf{24.9} &28.8 &\textbf{24.6} &\textbf{24.8} &\textbf{27.4} \\
  
  \toprule
  \end{tabular}
  }
  \label{table:h36m}
\end{table*}

\subsubsection{Bone direction consistency loss}
As each joint is always penalized independently, it is prone to fall into local optimum because of ignoring the graph structure of the human body. Therefore, the bone direction consistency loss is designed to check the bone direction keeping consistency between the original poses and the transformed poses. We first compute the bone vector to represent the bone direction, $\mathbf{b^{k}} = {\overrightarrow{ok}_{1} - \overrightarrow{ok}_{2}} = {X}^{k_{1}} - {X}^{k_{2}}$. 
Then, the bone direction consistency loss is calculated as follows:
\begin{equation}
  \small
  L_{b}=
  \sum_{k} ( 1 - \frac{\mathbf{{b^{k}_{1}}} \cdot \mathbf{b^{k}_{1\leftarrow2}}}{\|\mathbf{{b^{k}_{1}}}\| \|\mathbf{b^{k}_{1\leftarrow2}}\|} ) + \sum_{k} ( 1 - \frac{\mathbf{{b^{k}_{2}}} \cdot \mathbf{b^{k}_{2\leftarrow1}}}{\|\mathbf{{b^{k}_{2}}}\| \|\mathbf{b^{k}_{2\leftarrow1}}\|}),
\end{equation}
where $k$ is the index of bone for human body, $k_{1}$ and $k_{2}$ are the joints of the $k$-th bone.

\subsection{Learning Objective}
By performing both the single-view optimization and cross-view optimization, we combine the introduced losses: a 2D reprojection loss $L_{r}$, a symmetry loss $L_{s}$, a 3D transform consistency loss $L_{t}$, and a bone direction consistency loss $L_{b}$, as the final weakly-supervised learning objective,
\begin{equation}
  L=\lambda_{r} L_{r}+\lambda_{s} L_{s}+\lambda_{t} L_{t}+\lambda_{b} L_{b},
\end{equation}
where $\lambda_{r}=1$, $\lambda_{s}=1$, $\lambda_{t}=1$ and $\lambda_{b}=0.1$ are weighting factors.

\section{Experiments}

\begin{table*}
  \centering
  \caption{
    Quantitative results of our model trained with ground truth 2D pose of HumanEva-I dataset and tested with different levels of additive Gaussian noise $\mathcal{N}(0, \sigma)$ ($\sigma$ is the standard deviation) added to the triangulation. 
    The evaluation is performed under protocol \#2. 
    `Tri.' denotes the average results for triangulation and `Avg.' denotes the average results for refinement. 
  }
  \setlength{\tabcolsep}{2.80mm} 

  \begin{tabular}{@{}l|ccc|ccc|ccc|cc@{}}
    \toprule
    & \multicolumn{3}{c}{Walk} & \multicolumn{3}{c}{Jog} & \multicolumn{3}{c}{Box} \\
    & S1 & S2 & S3 & S1 & S2 & S3 & S1 & S2 & S3 &Avg. &Tri.\\
    \midrule
    Pavllo \emph{et al.}~\cite{pavllo20193d}, GT / GT &10.3 &8.3 &19.9 &15.8 &9.8 &11.4 &18.6 &25.3 &24.2 &15.9 & - \\
    Ours, GT / GT &2.3 &2.2 &2.9 &2.3 &2.4 &3.8 &2.5 &1.3 &7.0 &3.0 &6.0\\
    Ours, GT / GT + $\mathcal{N}(0,5)$ &6.1 &6.1 &6.4 &6.1 &6.2 &7.1 &6.4 &5.4 &10.5 &6.7 &8.5\\
    Ours, GT / GT + $\mathcal{N}(0,10)$ &11.0 &11.1 &11.4 &11.1 &11.1 &11.9 &11.3 &10.5 &15.0 &11.6 &12.8\\
    Ours, GT / GT + $\mathcal{N}(0,15)$ &16.2 &16.2 &16.4 &16.2 &16.3 &16.9 &16.4 &15.6 &19.7 &16.7 &17.6\\
    Ours, GT / GT + $\mathcal{N}(0,20)$ &21.3 &21.3 &21.6 &21.3 &21.4 &22.0 &21.4 &21.0 &24.4 &21.7 &22.5\\
    \bottomrule
  \end{tabular}

  \label{table:humaneva_eval}
\end{table*} 

\subsection{Datasets and Evaluation Metrics}

We evaluate our method on two standard benchmark datasets, Human3.6M~\cite{ionescu2013human3} and HumanEva-I~\cite{sigal2010humaneva}. 

\noindent \textbf{Human3.6M.}
The Human3.6M dataset is the largest publicly available benchmark dataset for 3D human pose estimation.
It consists of 3.6 million images captured from four synchronized 50 Hz cameras. 
There are 7 professional subjects performing 15 everyday activities. 
Following the standard protocol in prior work~\cite{chen2019weakly,tome2018rethinking}, we use 5 subjects (S1, S5, S6, S7, S8) for training and 2 subjects (S9 and S11) for evaluation. 

\noindent \textbf{HumanEva-I.}
HumanEva-I is a smaller dataset with fewer subjects and actions compared to Human3.6M, containing 3 subjects recorded from three synchronized camera views at 60 Hz. Following the train/test split in~\cite{pavllo20193d,lee2018propagating}, we train a single model on all subjects for all actions and test on validation sequences.

\noindent \textbf{Evaluation Metrics.}
We report the Mean Per Joint Position Error (MPJPE) and Procrustes analysis MPJPE (P-MPJPE) to measure the 3D pose estimation accuracy. 
MPJPE is the evaluation metric referred to as protocol \#1 in many works~\cite{fang2018learning,kocabas2019self}, which calculates the average Euclidean distance between the ground truth and predictions. 
P-MPJPE reports the error after the estimated 3D poses aligned to the ground truth in translation, rotation, and scale, which is referred to as protocol \#2~\cite{martinez2017simple,rayat2018exploiting}.

\subsection{Implementation Details}
In this work, all experiments are conducted on the PyTorch framework with one NVIDIA RTX 2080 Ti GPU.
Our model is trained using Amsgrad optimizer with a mini-batch size of 256 for Human3.6M. For HumanEva-I, the mini-batch size is set to 64.
An initial learning rate of 0.001 is used and decreases by 0.9 whenever the training loss does not decrease for every 10 epochs. 

Different from other multi-view methods that use all the camera views provided by the dataset, we only utilize two camera views for training and inference due to the consideration of computation complexity and implementing flexibility in the wild. For Human3.6M, adjacent camera pairs $P_1$ ($c_1$ and $c_2$), $P_2$ ($c_1$ and $c_3$), $P_3$ ($c_2$ and $c_4$), and $P_4$ ($c_3$ and $c_4$) are used for training and testing. 
As for HumanEva-I, adjacent camera pairs $\tilde{P_1}$ ($c_1$ and $c_2$), $\tilde{P_2}$ ($c_1$ and $c_3$), and $\tilde{P_3}$ ($c_2$ and $c_3$) are used. 

The 2D poses can be obtained by performing any classic 2D detection methods or directly using the 2D ground truth. Following~\cite{pavllo20193d}, we utilize the cascaded pyramid network (CPN)~\cite{chen2018cascaded} to obtain 2D poses of the Human3.6M dataset for a fair comparison. 
For HumanEva-I, we directly use the 2D ground truth to perform triangulation and then add Gaussian noise with the different variances to the triangulated 3D poses to validate the robustness of CV-UGCN.

\subsection{Comparison to the State-of-the-arts}
As shown in Table~\ref{table:h36m}, we report the MPJPE results of various methods on Human3.6M, monocular-based or multi-view based, using 3D ground truth or not, for a complete comparison. Monocular-based methods only use single-view information to estimate the poses, suffering from the depth ambiguity problem. To overcome this inherent limitation, some methods~\cite{pavllo20193d, chen2021anatomy} exploit temporal information in monocular video sequences and achieve better performance. By making full use of the multi-view information and the geometry knowledge, multi-view based methods are more applicable. It can be seen that the performance of multi-view based methods is obviously superior to monocular-based methods, even when no 3D labels are utilized for training. Compared with these methods, our method makes a further improvement and gains a new state-of-the-art performance of 27.4 mm, benefitting from our effective refinement model and weakly-supervised learning objective. 

\subsection{Robustness Evaluation}
\label{Sec:robustness}
To evaluate the robustness of CV-UGCN, experiments are conducted on HumanEva-I dataset to test the tolerance for triangulation noises. To this end, we use 2D ground-truth poses for triangulation and train a single model following the train/test split in~\cite{pavllo20193d}. Then, triangulated poses added with the Gaussian noise of different variances are fed into the CV-UGCN for refinement. The results are shown in Table~\ref{table:humaneva_eval}.

Although the 3D poses are triangulated from 2D ground truth, it still exists 6.0 mm P-MPJPE due to the camera calibration error. Our refinement model can refine the results to 3.0 mm, demonstrating the effectiveness of CV-UGCN. Besides, when the triangulated poses are added with Gaussian noises of different variances, the system seems not to be broken, with a stable improvement of around 1.0 mm compared to the input coarse poses. It validates that our refinement model can perform reasonably and is robust to the coarse 3D poses with different levels of noise.

\begin{table}[!t]
  \centering
  \caption
  {
      Ablation studies on each component of CV-UGCN. 
      The evaluation is performed on Human3.6M with MPJPE metric under protocol \#1. $\Delta$ represents the performance gap between the methods and Ours (CV-UGCN).
  }
  \setlength{\tabcolsep}{3.80mm} 

  \begin{tabular}{lcc}
    \toprule  
      Method& MPJPE (mm) & $\Delta$ \\
      \midrule  
  w/o refinement & 36.0 &8.6 \\ 
      \midrule  
      Fully-connected  network &31.8 &4.4 \\
      Ours w/o spatial configuration & 31.5 &4.1 \\
      Ours w/o cross-view correlation & 29.3 &1.9 \\
      Ours w/o fusion & 28.2 &0.8 \\
      Ours (CV-UGCN)  & 27.4 &- \\
      \bottomrule 
  \end{tabular}

  \label{table:ablation_method}
\end{table}

\begin{table}[!t]
  \centering
  \caption
  {
    Ablation studies on each term of the weakly-supervised losses. $\Delta$ represents the performance gap between the methods and the last-row method.
  }
  \setlength{\tabcolsep}{6.70mm} 
  
  \begin{tabular}{lcc}
  \toprule  
  Method& MPJPE (mm)& $\Delta$  \\
  \midrule  
  $L_{r}$ & 28.3 & 0.9\\
  $L_{r} + L_{s}$ & 28.2& 0.8 \\
  $L_{r} + L_{t}$ & 28.1& 0.7 \\
  $L_{r} + L_{b}$ & 27.9& 0.5 \\
  $L_{r} + L_{s} + L_{b}$ & 27.8& 0.4 \\
  $L_{r} + L_{s} + L_{b} + L_{t}$ & 27.4 & -\\
  \bottomrule 
  \end{tabular}

  \label{table:ablation_loss}
\end{table}

\subsection{Ablation Studies}
Sufficient ablation studies are performed on Human3.6M dataset under protocol \#1 to verify the effectiveness of our method.

\subsubsection{Effect of CV-UGCN}
An ablation study is performed by changing various components of CV-UGCN to show the contribution of each component of our refinement model, as presented in Table~\ref{table:ablation_method}. 
\begin{itemize}
  \item \textbf{w/o refinement}: Using triangulation to obtain 3D poses from 2D detections.
  
  \item \textbf{Fully-connected network}: We utilize the fully-connected network like~\cite{martinez2017simple} as the refinement model to refine the coarse triangulated 3D poses.

  \item \textbf{Ours w/o spatial configuration}: We omit the spatial configuration from the graph model of CV-UGCN.
  
  \item \textbf{Ours w/o cross-view correlation}: We omit the cross-view correlation from the graph model of CV-UGCN.
  
  \item \textbf{Ours w/o fusion}: The cross-view triangulated 3D poses are not first preprocessed by S-GCN units to get the fused cross-view features. Instead, they are directly concatenated and sent to the M-GCN units.
  
  \item \textbf{Ours (CV-UGCN)}: The full model of our proposed method.
\end{itemize}

Without exploiting the refinement model, the triangulated results are noisy with the MPJPE of 36.0 mm. When using the fully-connected network as the refinement model, the results are improved to 31.8 mm. 
CV-UGCN can refine the coarse 3D poses more effectively and bring significant improvement (8.6 mm). 
When removing the spatial configuration or cross-view correlation from the adjacency matrix of CV-UGCN, the performance decreases. 
It verifies the importance of considering both the spatial configuration and cross-view correlation in the graph model. 
Meanwhile, without first using S-GCNs to preprocess the coarse poses of each view and then fusing the features for M-GCNs, the result (ours w/o fusion) also decreases. 
This is because the triangulated poses are too noisy to conduct cross-view information interaction well in the early stage. 

\begin{figure*}[tb]
  \centering
  \subfigure[]{\includegraphics[width=0.495\linewidth]{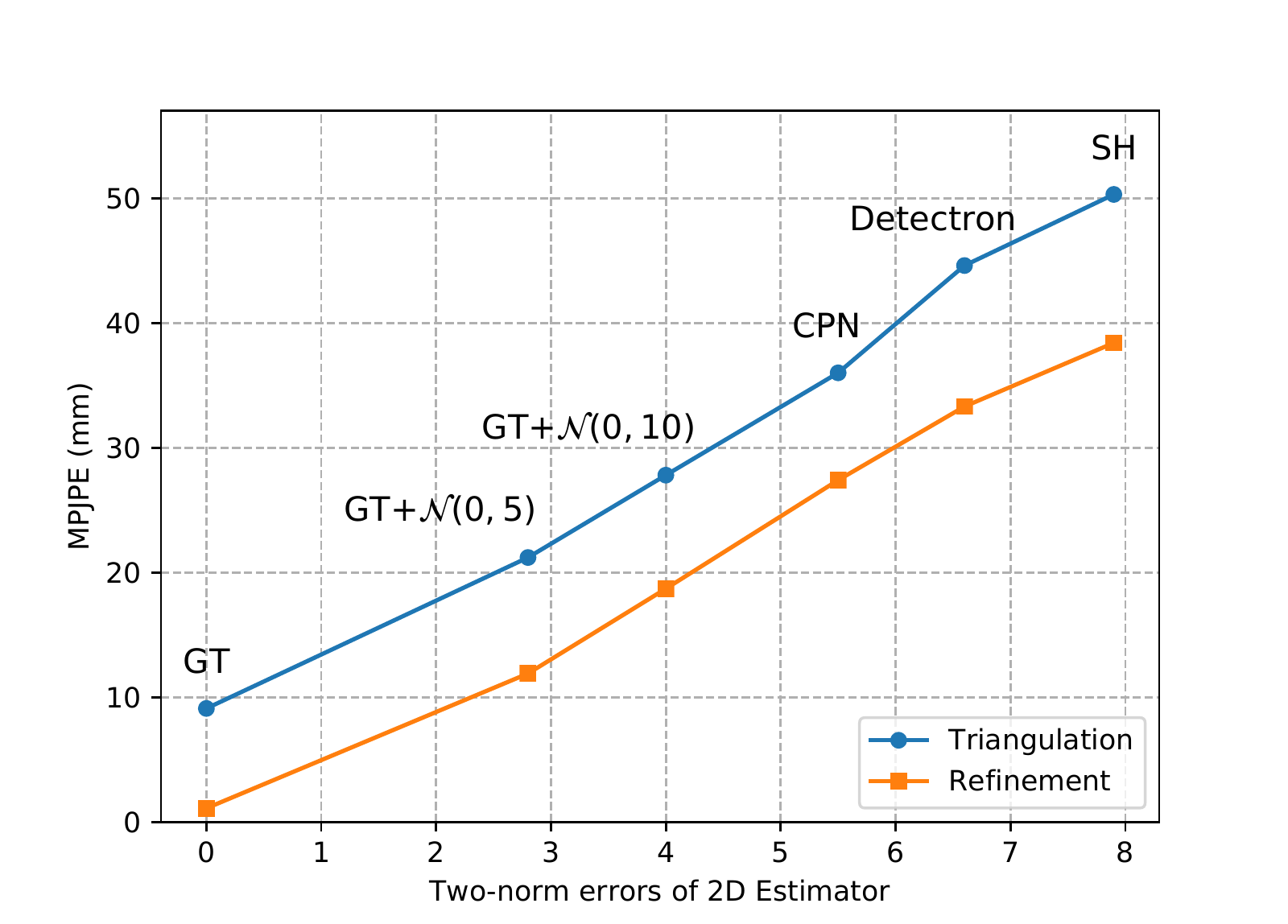}}
  \subfigure[]{\includegraphics[width=0.495\linewidth]{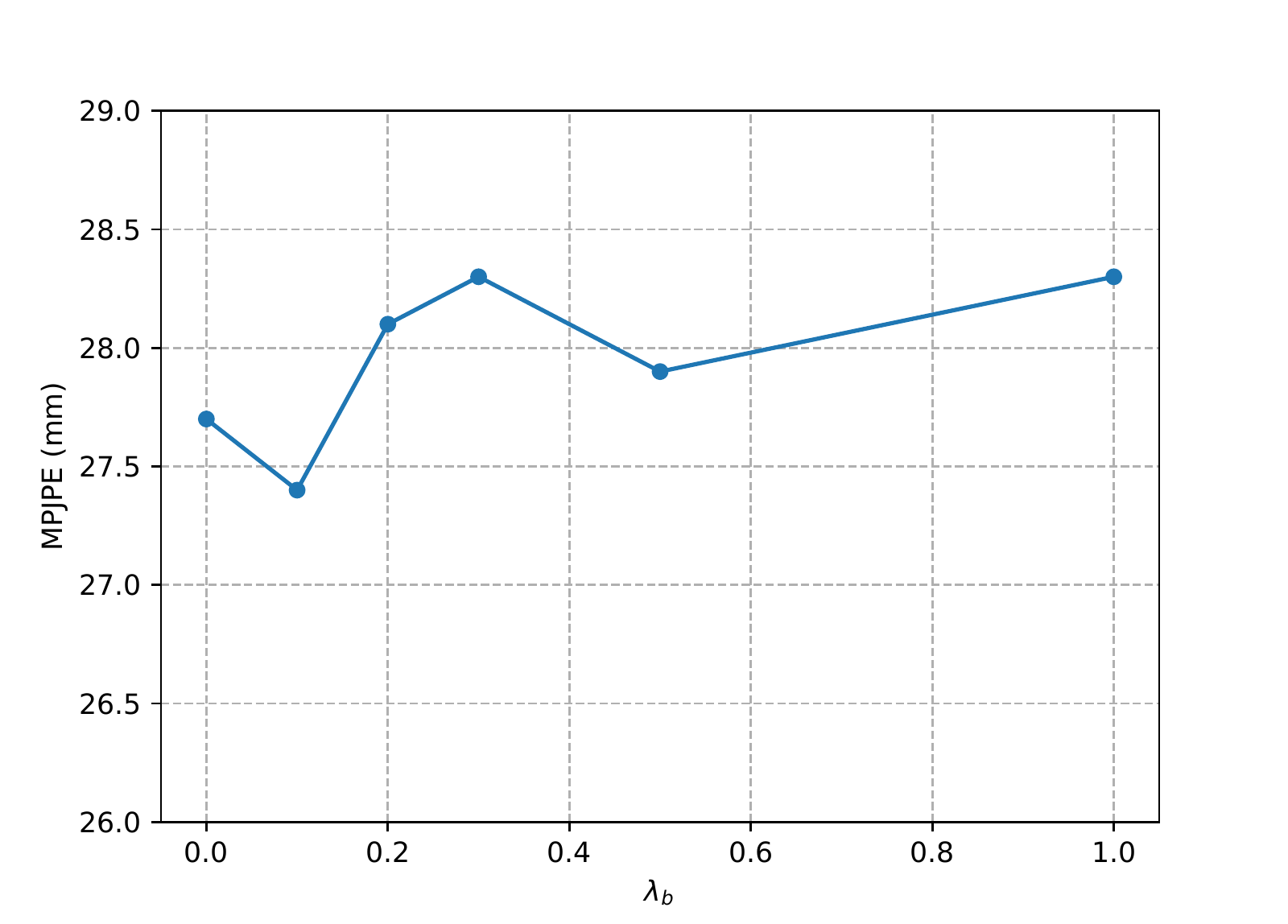}}
  \caption
  {
    (a) The impact of 2D detections. Experiments are performed on Human3.6M with the MPJPE metric under the protocol \#1. $\mathcal{N}(0, \sigma^2)$ represents the Gaussian noise with mean zero and standard deviation $\sigma$. GT denotes the 2D annotations.
    (b) Evaluation of various weighting factors $\lambda_b$ for bone direction consistency loss $L_b$.
    Experiments are conducted on Human3.6M with the MPJPE metric.
  }
  \label{fig:2D and loss}
\end{figure*}

\begin{table*}[!t]
  \centering
  \caption
  {
  MPJPE Results (in mm) of our method on different camera paired views on Human3.6M. 
  We evaluate the generalization capabilities of our approach testing on unseen camera paired views. 
  $P_{i}$ denotes $i$-th camera paired views. 
  The results of the unseen paired views are marked in \textbf{bold}.
  }
  \setlength{\tabcolsep}{7.40mm} 
  
  \begin{tabular}{l|cccc|c}
  \toprule  
  Method / Test camera views & $P_1$& $P_2$& $P_3$& $P_4$& Avg.\\
  \midrule  
  Triangulation & 35.8& 40.1& 32.6& 35.4& 36.0 \\ \midrule
  Training on $P_3$& \textbf{35.4}& \textbf{35.9}& 26.5& \textbf{35.2}& 33.3 \\
  Training on $P_1$ and $P_3$& 28.2& \textbf{35.7}& 25.9& \textbf{31.7}& 30.4 \\
  Training on $P_2$ and $P_4$& \textbf{32.6}& 30.7& \textbf{30.4}& 27.1& 30.2 \\ 
  Training on $P_1$, $P_2$, $P_3$& 28.1& 30.7& 26.0& \textbf{31.1} &29.0 \\
  Training on $P_1$, $P_2$, $P_4$& 27.9& 30.9& \textbf{30.2}& 27.2& 29.1 \\
  Training on all pairs& 27.4& 30.3& 25.3& 26.8& 27.4 \\
  \bottomrule 
  \end{tabular}

  \label{table:camera}
  \end{table*}
 
 \begin{table*}[!t]
  \centering
  \caption
  {
  Model size and performance comparison. 
  3D means that these methods use 3D annotations to train the model.
  $V$ is the number of camera views for training and inference. 
  2D-GT + Ours represents using 2D ground truth as input.
  }
  \setlength{\tabcolsep}{6.7mm} 
  
  \begin{tabular}{l|cccc}
  \toprule  
  Method& $V$& 3D& Model Size& MPJPE (mm)  \\
  \midrule  
  Qiu \emph{et al.} (CVPR'19)~\cite{qiu2019cross} &4 & & 2.1 GB & 31.2\\
  Iskakov \emph{et al.} (ICCV'19)~\cite{iskakov2019learnable} &4 &\checkmark & 643 MB& 20.8 \\
  Remelli \emph{et al.} (CVPR'20)~\cite{remelli2020lightweight} &4 &\checkmark & 251 MB& 30.2 \\
  CPN + Ours &2 & & 47 MB& 27.4 \\
  \midrule  
  2D-GT + Ours &2 & & 20 MB& 1.1 \\
  \bottomrule 
  \end{tabular}

  \label{table:modelsize}
  \end{table*}

  \begin{figure*}[!t]
    \centering
    \includegraphics[width=1.0 \linewidth]{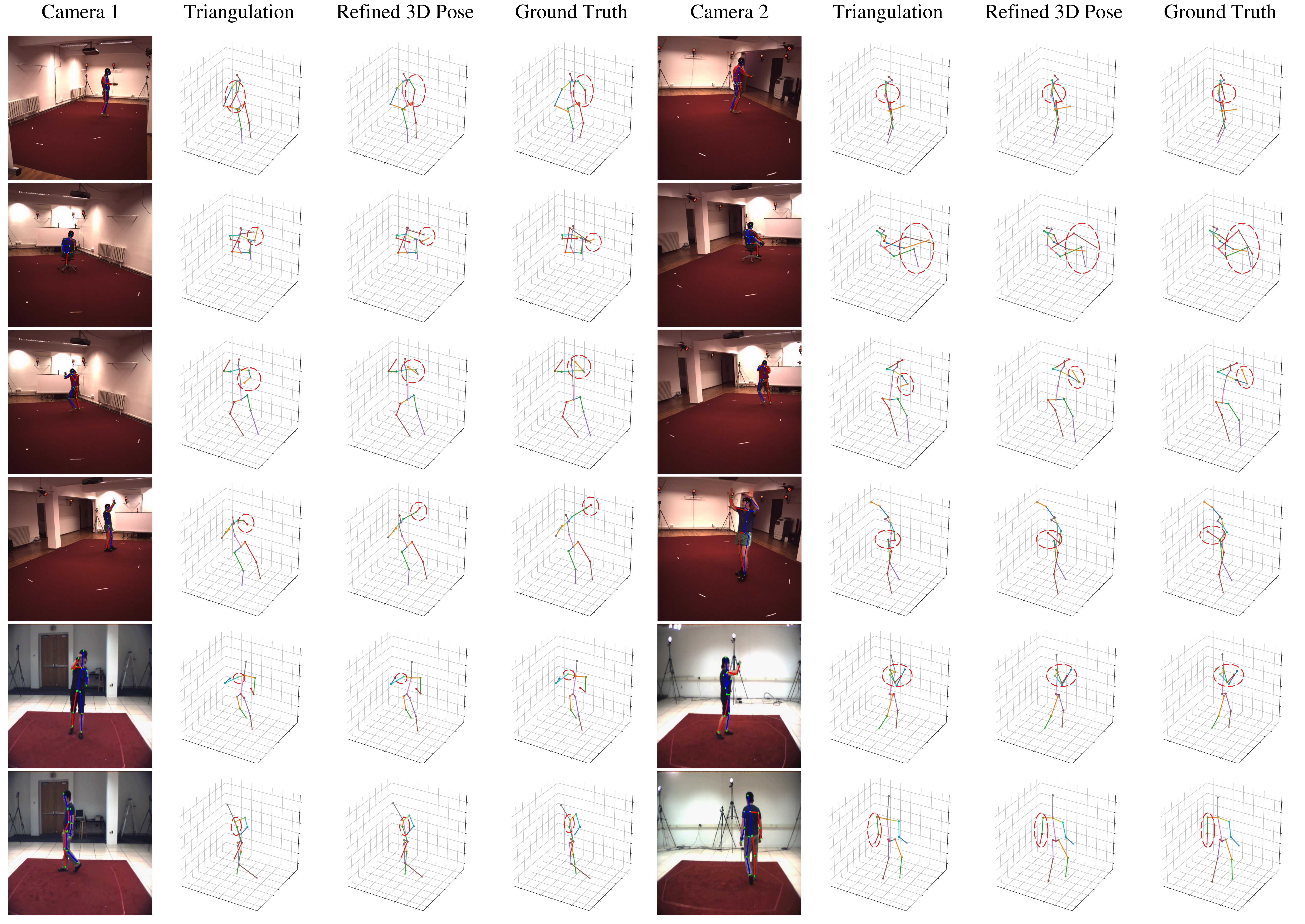}
    \caption
    {
    Qualitative results of our approach on Human3.6M dataset (first 3 rows) and HumanEva-I dataset (last 2 rows). 
    The results of triangulation are noisy and unreliable, while our model is able to produce realistic and structurally plausible 3D poses. 
    }
    \label{fig:results}
  \end{figure*}

\subsubsection{Effect of weakly-supervised losses}
In Table~\ref{table:ablation_loss}, we investigate the effect of each component of our weakly-supervised learning objective.
When only using the reprojection loss $L_r$, the MPJPE is 28.3 mm. If $L_r$ works with an additional loss (symmetry loss $L_{s}$, 3D transform consistency loss $L_{t}$ or bone direction consistency loss $L_{b}$), the performance is improved. At last, as more components are used, the results are better. This is attributed to the joint geometric and structure-aware consistency checks that are performed in both single-view and cross-view optimization manners.
    
  \begin{figure*}[!t]
  \centering
  \includegraphics[width=0.6 \linewidth]{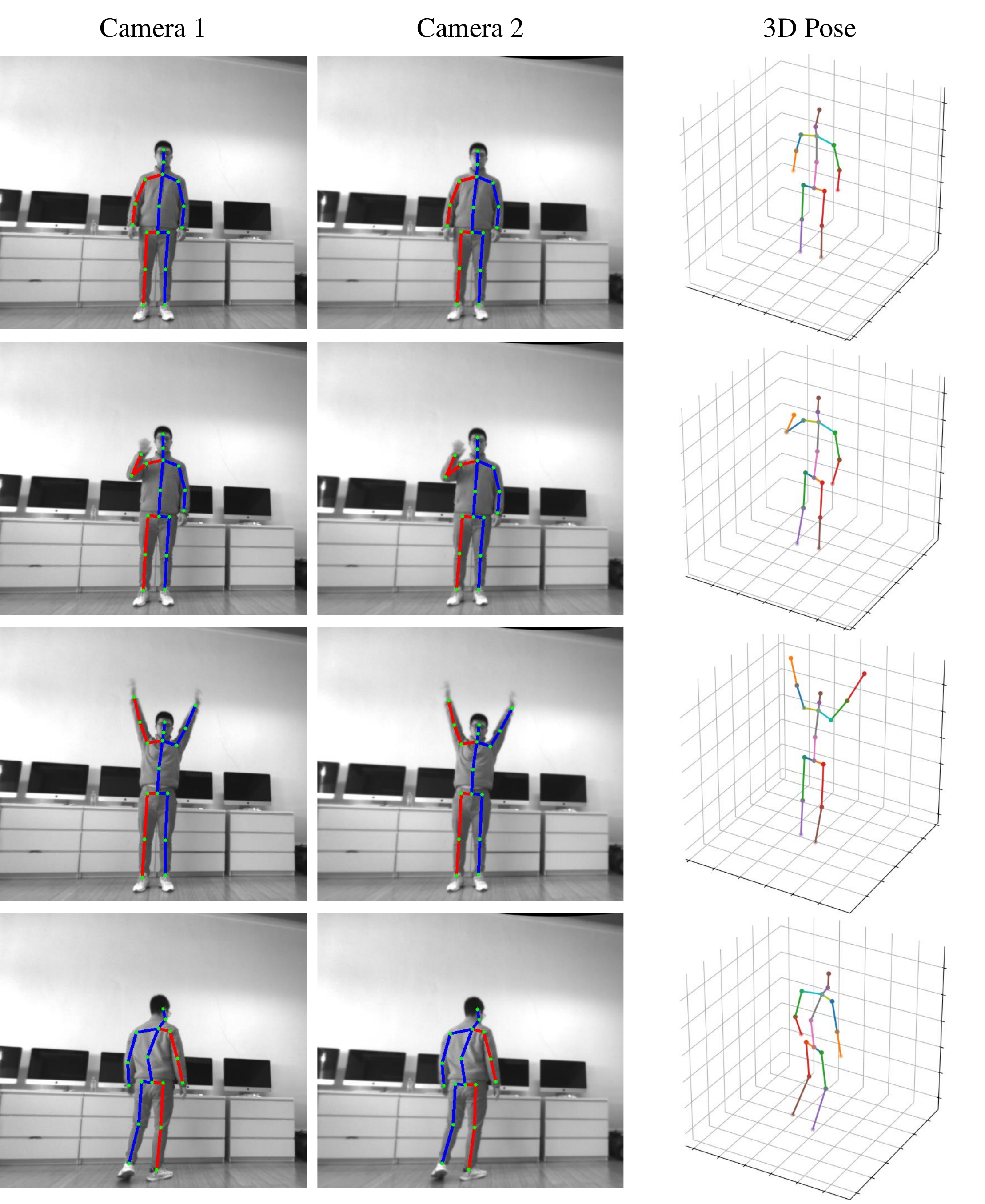}
  \caption
  {
  Visualization results of our method in the real scene. Camera 1 and Camera 2 are the left and right views of a stereo camera.
  }
  \label{fig:wild}
  \end{figure*}

\subsubsection{Impact of different 2D detections}
\label{Sec:2D evaluation}
For the 2D-3D lifting task, the precision of 2D detections is important to the performance of estimating 3D poses~\cite{martinez2017simple}. We utilize Stack Hourglass (SH)~\cite{newell2016stacked}, Detectron~\cite{pavllo20193d}, CPN~\cite{chen2018cascaded}, and 2D ground truth (GT) with different levels of additive Gaussian noises to explore the impact of different 2D detections. Figure~\ref{fig:2D and loss} (a) shows the relationship between the MPJPE of 3D poses and two-norm errors of 2D detections. For both the triangulation and the refinement, the MPJPE of 3D poses increases linearly with the two-norm errors of 2D detections. However, it can be observed that our refinement model has a lower incremental rate than triangulation, verifying the effectiveness of our refinement model to refine the triangulated 3D poses. Meanwhile, when the 2D ground truth is directly used for triangulation, the results still have 9.1 mm MPJPE compared to the 3D ground truth. It shows that the triangulation is sensitive to the camera calibration noises. The proposed refinement model can effectively refine the results to 1.1 mm MPJPE.

\subsubsection{Hyperparameter evaluations}
\label{Sec:wb evaluation}
Due to the bone direction consistency loss is calculated by measuring the cosine distance between two bone vectors which is different from other losses that calculate the Mean Squared Error (MSE), it is necessary to find a proper weighting factor for $L_b$. As shown in Figure~\ref{fig:2D and loss} (b), the best performance is achieved when $\lambda_b$ is equal to 0.1.

\subsection{Generalization to Unseen Cameras}
\label{Sec:unseen camera}
To validate the feasibility of applying our model to unknown views, we train the model on some paired views and evaluate its performance on other unseen camera pairs. 
The results are presented in Table~\ref{table:camera}, where $P_{i}$ denotes $i$-th paired views. 
The results on unseen paired camera views are marked in bold. It can be observed that CV-UGCN still can improve the triangulated results in unseen camera views. For example, the model trained only on $P_3$ can also effectively refine the triangulation results on $P_1$, $P_2$ and $P_4$. It verifies the generalization capability to unseen camera views of our method. 

\subsection{Model Size and Inference Time}
We report the model size and inference time to show the efficiency of our methods. Table~\ref{table:modelsize} exhibits the comparison of the model size and performance with the recent methods~\cite{qiu2019cross,iskakov2019learnable,remelli2020lightweight}. Because the comparison methods embed the 2D detector into their model, we add the model size of CPN~\cite{chen2018cascaded} to ours for a fair comparison. It can be seen that our method can achieve impressive performance with a lightweight model.

On a machine equipped with one NVIDIA RTX 2080 Ti GPU, the 2D detector CPN requires about 0.02s to perform the 2D detection, while our pipeline needs about another 0.01s to estimate the 3D poses. Consequently, when estimating 3D poses from images of two camera views, our method could yield a real-time performance ($\sim$33 fps). Note that if a better and faster 2D detector is used with our method, the performance and speed can have a further improvement.

\subsection{Qualitative Results}
\noindent \textbf{3D reconstruction visualization. }
\label{Sec:visual}
Figure~\ref{fig:results} shows some visualization results of the triangulation and refinement model. The first 3 rows are the results on Human3.6M dataset and the last 2 rows are on HumanEva-I dataset. 
Because of the error of 2D detections and camera calibration errors, the triangulation results are noisy and unreliable (see the poses in the red dotted circles). 
Our refinement model can efficiently improve the coarse 3D poses to be more realistic and structurally plausible.

\noindent \textbf{Visual results in the Real Scene. }
\label{Sec:real scene}
We apply our method to the real scene to test its applicability. By using a stereo camera, we take photos in a room and use the proposed method to estimate the absolute 3D poses of the actor. Some visualization results are given in Figure~\ref{fig:wild}. It can be seen that the results are reasonable, showing the feasibility of our method in real-scene applications. 

\section{Conclusion}
In this paper, a simple yet effective pipeline is proposed for cross-view 3D human pose estimation in a coarse-to-fine manner. Specifically, we first exploit triangulation to lift the 2D detections to coarse 3D poses and then utilize a refinement model to obtain precise results. In particular, a new cross-view U-shaped graph convolutional network (CV-UGCN) is designed as the refinement model, which can take advantage of spatial configurations and cross-view correlations to accurately refine the coarse 3D poses. Moreover, to release the requirement of quantities of 3D ground truth as supervision, we introduce a weakly-supervised learning objective by exploiting geometric and structure-aware consistency checks in both single-view and cross-view optimizations. Extensive experiments have been conducted on the benchmark dataset. The results show that our method not only achieves state-of-the-art performance but also is lightweight and could run in real-time.  

\bibliographystyle{IEEEtran}
\bibliography{ref.bib}

\begin{thebibliography}{10}
\providecommand{\url}[1]{#1}
\csname url@samestyle\endcsname
\providecommand{\newblock}{\relax}
\providecommand{\bibinfo}[2]{#2}
\providecommand{\BIBentrySTDinterwordspacing}{\spaceskip=0pt\relax}
\providecommand{\BIBentryALTinterwordstretchfactor}{4}
\providecommand{\BIBentryALTinterwordspacing}{\spaceskip=\fontdimen2\font plus
\BIBentryALTinterwordstretchfactor\fontdimen3\font minus
  \fontdimen4\font\relax}
\providecommand{\BIBforeignlanguage}[2]{{%
\expandafter\ifx\csname l@#1\endcsname\relax
\typeout{** WARNING: IEEEtran.bst: No hyphenation pattern has been}%
\typeout{** loaded for the language `#1'. Using the pattern for}%
\typeout{** the default language instead.}%
\else
\language=\csname l@#1\endcsname
\fi
#2}}
\providecommand{\BIBdecl}{\relax}
\BIBdecl

\bibitem{li20143d}
S.~Li and A.~B. Chan, ``{3D} human pose estimation from monocular images with
  deep convolutional neural network,'' in \emph{Proceedings of the Asian
  Conference on Computer Vision (ACCV)}, 2014, pp. 332--347.

\bibitem{chen20173d}
C.-H. Chen and D.~Ramanan, ``{3D} human pose estimation = {2D} pose estimation
  + matching,'' in \emph{Proceedings of the IEEE Conference on Computer Vision
  and Pattern Recognition (CVPR)}, 2017, pp. 7035--7043.

\bibitem{zhou2017towards}
X.~Zhou, Q.~Huang, X.~Sun, X.~Xue, and Y.~Wei, ``Towards {3D} human pose
  estimation in the wild: a weakly-supervised approach,'' in \emph{Proceedings
  of the IEEE International Conference on Computer Vision (ICCV)}, 2017, pp.
  398--407.

\bibitem{martinez2017simple}
J.~Martinez, R.~Hossain, J.~Romero, and J.~J. Little, ``A simple yet effective
  baseline for {3D} human pose estimation,'' in \emph{Proceedings of the IEEE
  International Conference on Computer Vision (ICCV)}, 2017, pp. 2640--2649.

\bibitem{fabbri2020compressed}
M.~Fabbri, F.~Lanzi, S.~Calderara, S.~Alletto, and R.~Cucchiara, ``Compressed
  volumetric heatmaps for multi-person {3D} pose estimation,'' in
  \emph{Proceedings of the IEEE Conference on Computer Vision and Pattern
  Recognition (CVPR)}, 2020, pp. 7204--7213.

\bibitem{strided}
W.~Li, H.~Liu, R.~Ding, M.~Liu, P.~Wang, and W.~Yang, ``Exploiting temporal
  contexts with strided transformer for {3D} human pose estimation,''
  \emph{IEEE Transactions on Multimedia}, 2022.

\bibitem{mhformer}
W.~Li, H.~Liu, H.~Tang, P.~Wang, and L.~Van~Gool, ``{MHF}ormer:
  Multi-hypothesis transformer for {3D} human pose estimation,'' in
  \emph{Proceedings of the IEEE Conference on Computer Vision and Pattern
  Recognition (CVPR)}, 2022.

\bibitem{harvesting2017}
G.~Pavlakos, X.~Zhou, K.~G. Derpanis, and K.~Daniilidis, ``Harvesting multiple
  views for marker-less {3D} human pose annotations,'' in \emph{Proceedings of
  the IEEE Conference on Computer Vision and Pattern Recognition (CVPR)}, 2017,
  pp. 6988--6997.

\bibitem{iskakov2019learnable}
K.~Iskakov, E.~Burkov, V.~Lempitsky, and Y.~Malkov, ``Learnable triangulation
  of human pose,'' in \emph{Proceedings of the IEEE International Conference on
  Computer Vision (ICCV)}, 2019, pp. 7718--7727.

\bibitem{qiu2019cross}
H.~Qiu, C.~Wang, J.~Wang, N.~Wang, and W.~Zeng, ``Cross view fusion for {3D}
  human pose estimation,'' in \emph{Proceedings of the IEEE International
  Conference on Computer Vision (ICCV)}, 2019, pp. 4342--4351.

\bibitem{kocabas2019self}
M.~Kocabas, S.~Karagoz, and E.~Akbas, ``Self-supervised learning of {3D} human
  pose using multi-view geometry,'' in \emph{Proceedings of the IEEE Conference
  on Computer Vision and Pattern Recognition (CVPR)}, 2019, pp. 1077--1086.

\bibitem{chen2014articulated}
X.~Chen and A.~L. Yuille, ``Articulated pose estimation by a graphical model
  with image dependent pairwise relations,'' in \emph{Advances in Neural
  Information Processing Systems (NeurIPS)}, 2014, pp. 1736--1744.

\bibitem{pavlakos2017coarse}
G.~Pavlakos, X.~Zhou, K.~G. Derpanis, and K.~Daniilidis, ``Coarse-to-fine
  volumetric prediction for single-image {3D} human pose,'' in
  \emph{Proceedings of the IEEE Conference on Computer Vision and Pattern
  Recognition (CVPR)}, 2017, pp. 7025--7034.

\bibitem{cai2019exploiting}
Y.~Cai, L.~Ge, J.~Liu, J.~Cai, T.-J. Cham, J.~Yuan, and N.~M. Thalmann,
  ``Exploiting spatial-temporal relationships for {3D} pose estimation via
  graph convolutional networks,'' in \emph{Proceedings of the IEEE
  International Conference on Computer Vision (ICCV)}, 2019, pp. 2272--2281.

\bibitem{remelli2020lightweight}
E.~Remelli, S.~Han, S.~Honari, P.~Fua, and R.~Wang, ``Lightweight multi-view
  {3D} pose estimation through camera-disentangled representation,'' in
  \emph{Proceedings of the IEEE Conference on Computer Vision and Pattern
  Recognition (CVPR)}, 2020, pp. 6040--6049.

\bibitem{wandt2019repnet}
B.~Wandt and B.~Rosenhahn, ``{R}ep{N}et: Weakly supervised training of an
  adversarial reprojection network for {3D} human pose estimation,'' in
  \emph{Proceedings of the IEEE Conference on Computer Vision and Pattern
  Recognition (CVPR)}, 2019, pp. 7782--7791.

\bibitem{kundu2020self}
J.~N. Kundu, S.~Seth, V.~Jampani, M.~Rakesh, R.~V. Babu, and A.~Chakraborty,
  ``Self-supervised {3D} human pose estimation via part guided novel image
  synthesis,'' in \emph{Proceedings of the IEEE Conference on Computer Vision
  and Pattern Recognition (CVPR)}, 2020, pp. 6152--6162.

\bibitem{kipf2016semi}
T.~N. Kipf and M.~Welling, ``Semi-supervised classification with graph
  convolutional networks,'' \emph{arXiv preprint arXiv:1609.02907}, 2016.

\bibitem{ronneberger2015u}
O.~Ronneberger, P.~Fischer, and T.~Brox, ``U-{N}et: Convolutional networks for
  biomedical image segmentation,'' in \emph{Proceedings of the International
  Conference on Medical Image Computing and Computer-assisted Intervention},
  2015, pp. 234--241.

\bibitem{wang2020motion}
J.~Wang, S.~Yan, Y.~Xiong, and D.~Lin, ``Motion guided {3D} pose estimation
  from videos,'' in \emph{Proceedings of the European Conference on Computer
  Vision (ECCV)}, 2020, pp. 764--780.

\bibitem{dabral2018learning}
R.~Dabral, A.~Mundhada, U.~Kusupati, S.~Afaque, A.~Sharma, and A.~Jain,
  ``Learning {3D} human pose from structure and motion,'' in \emph{Proceedings
  of the European Conference on Computer Vision (ECCV)}, 2018, pp. 668--683.

\bibitem{tome2017lifting}
D.~Tome, C.~Russell, and L.~Agapito, ``Lifting from the deep: Convolutional
  {3D} pose estimation from a single image,'' in \emph{Proceedings of the IEEE
  Conference on Computer Vision and Pattern Recognition (CVPR)}, 2017, pp.
  2500--2509.

\bibitem{wang2019distill}
C.~Wang, C.~Kong, and S.~Lucey, ``Distill knowledge from nrsfm for weakly
  supervised {3D} pose learning,'' in \emph{Proceedings of the IEEE
  International Conference on Computer Vision (ICCV)}, 2019, pp. 743--752.

\bibitem{xu2021graph}
T.~Xu and W.~Takano, ``Graph stacked hourglass networks for {3D} human pose
  estimation,'' in \emph{Proceedings of the IEEE Conference on Computer Vision
  and Pattern Recognition (CVPR)}, 2021, pp. 16\,105--16\,114.

\bibitem{pavllo20193d}
D.~Pavllo, C.~Feichtenhofer, D.~Grangier, and M.~Auli, ``{3D} human pose
  estimation in video with temporal convolutions and semi-supervised
  training,'' in \emph{Proceedings of the IEEE Conference on Computer Vision
  and Pattern Recognition (CVPR)}, 2019, pp. 7753--7762.

\bibitem{liu2020attention}
R.~Liu, J.~Shen, H.~Wang, C.~Chen, S.-c. Cheung, and V.~Asari, ``Attention
  mechanism exploits temporal contexts: Real-time {3D} human pose
  reconstruction,'' in \emph{Proceedings of the IEEE Conference on Computer
  Vision and Pattern Recognition (CVPR)}, 2020, pp. 5064--5073.

\bibitem{zeng2020srnet}
A.~Zeng, X.~Sun, F.~Huang, M.~Liu, Q.~Xu, and S.~Lin, ``{SRN}et: Improving
  generalization in {3D} human pose estimation with a split-and-recombine
  approach,'' in \emph{Proceedings of the European Conference on Computer
  Vision (ECCV)}, 2020, pp. 507--523.

\bibitem{chen2021anatomy}
T.~Chen, C.~Fang, X.~Shen, Y.~Zhu, Z.~Chen, and J.~Luo, ``Anatomy-aware 3d
  human pose estimation with bone-based pose decomposition,'' \emph{IEEE
  Transactions on Circuits and Systems for Video Technology}, vol.~32, no.~1,
  pp. 198--209, 2021.

\bibitem{trumble2017total}
M.~Trumble, A.~Gilbert, C.~Malleson, A.~Hilton, and J.~Collomosse, ``Total
  capture: {3D} human pose estimation fusing video and inertial sensors,'' in
  \emph{Proceedings of the British Machine Vision Conference (BMVC)}, 2017, pp.
  1--13.

\bibitem{iqbal2020weakly}
U.~Iqbal, P.~Molchanov, and J.~Kautz, ``Weakly-supervised {3D} human pose
  learning via multi-view images in the wild,'' in \emph{Proceedings of the
  IEEE Conference on Computer Vision and Pattern Recognition (CVPR)}, 2020, pp.
  5243--5252.

\bibitem{tome2018rethinking}
D.~Tome, M.~Toso, L.~Agapito, and C.~Russell, ``Rethinking pose in {3D}:
  Multi-stage refinement and recovery for markerless motion capture,'' in
  \emph{Proceedings of the International Conference on {3D} Vision (3DV)},
  2018, pp. 474--483.

\bibitem{chen2019weakly}
X.~Chen, K.-Y. Lin, W.~Liu, C.~Qian, and L.~Lin, ``Weakly-supervised discovery
  of geometry-aware representation for {3D} human pose estimation,'' in
  \emph{Proceedings of the IEEE Conference on Computer Vision and Pattern
  Recognition (CVPR)}, 2019, pp. 10\,895--10\,904.

\bibitem{he2020epipolar}
Y.~He, R.~Yan, K.~Fragkiadaki, and S.-I. Yu, ``Epipolar transformers,'' in
  \emph{Proceedings of the IEEE Conference on Computer Vision and Pattern
  Recognition (CVPR)}, 2020, pp. 7779--7788.

\bibitem{ionescu2013human3}
C.~Ionescu, D.~Papava, V.~Olaru, and C.~Sminchisescu, ``Human3.6{M}: Large
  scale datasets and predictive methods for {3D} human sensing in natural
  environments,'' \emph{IEEE Transactions on Pattern Analysis and Machine
  Intelligence}, vol.~36, no.~7, pp. 1325--1339, 2013.

\bibitem{sigal2010humaneva}
L.~Sigal, A.~O. Balan, and M.~J. Black, ``{H}uman{E}va: Synchronized video and
  motion capture dataset and baseline algorithm for evaluation of articulated
  human motion,'' \emph{International Journal of Computer Vision}, vol.~87, no.
  1-2, p.~4, 2010.

\bibitem{lee2018propagating}
K.~Lee, I.~Lee, and S.~Lee, ``Propagating {LSTM}: {3D} pose estimation based on
  joint interdependency,'' in \emph{Proceedings of the European Conference on
  Computer Vision (ECCV)}, 2018, pp. 119--135.

\bibitem{fang2018learning}
H.-S. Fang, Y.~Xu, W.~Wang, X.~Liu, and S.-C. Zhu, ``Learning pose grammar to
  encode human body configuration for {3D} pose estimation,'' in
  \emph{Proceedings of the AAAI Conference on Artificial Intelligence},
  vol.~32, no.~1, 2018.

\bibitem{rayat2018exploiting}
M.~Rayat Imtiaz~Hossain and J.~J. Little, ``Exploiting temporal information for
  {3D} human pose estimation,'' in \emph{Proceedings of the European Conference
  on Computer Vision (ECCV)}, 2018, pp. 68--84.

\bibitem{chen2018cascaded}
Y.~Chen, Z.~Wang, Y.~Peng, Z.~Zhang, G.~Yu, and J.~Sun, ``Cascaded pyramid
  network for multi-person pose estimation,'' in \emph{Proceedings of the IEEE
  Conference on Computer Vision and Pattern Recognition (CVPR)}, 2018, pp.
  7103--7112.

\bibitem{newell2016stacked}
A.~Newell, K.~Yang, and J.~Deng, ``Stacked hourglass networks for human pose
  estimation,'' in \emph{Proceedings of the European Conference on Computer
  Vision (ECCV)}, 2016, pp. 483--499.

\end{thebibliography}

\end{document}